\documentclass[letterpaper]{article} 
\usepackage{aaai25}  
\usepackage{times}  
\usepackage{helvet}  
\usepackage{courier}  
\usepackage[hyphens]{url}  
\usepackage{graphicx} 
\usepackage{natbib}  
\usepackage{caption} 
\PassOptionsToPackage{dvipsnames}{xcolor}
\usepackage{subcaption}
\usepackage{booktabs}
\usepackage{multirow}
\usepackage{float}
\usepackage{algorithm}
\usepackage{algorithmicx}
\usepackage{algpseudocode}
\usepackage{color}
\usepackage{xcolor}
\definecolor{Green}{rgb}{0,0.6,0}
\usepackage{amsmath}
\usepackage{amsfonts}

\usepackage{newfloat}
\usepackage{listings}
\DeclareCaptionStyle{ruled}{labelfont=normalfont,labelsep=colon,strut=off} 
\floatstyle{ruled}
\newfloat{listing}{tb}{lst}{}
\floatname{listing}{Listing}
\title{Filter or Compensate: Towards Invariant Representation from Distribution Shift for Anomaly Detection}
\author{
    Zining Chen\textsuperscript{\rm 1},
    Xingshuang Luo\textsuperscript{\rm 1},
    Weiqiu Wang\textsuperscript{\rm 1},
    Zhicheng Zhao\textsuperscript{\rm 1,2,3}\equalcontrib,
    Fei Su\textsuperscript{\rm 1,2,3},
    Aidong Men\textsuperscript{\rm 1}
}
\affiliations{
    \textsuperscript{\rm 1}School of Artificial Intelligence, Beijing University of Posts and Telecommunications\\
    \textsuperscript{\rm 2}Beijing Key Laboratory of Network System and Network Culture, China\\
    \textsuperscript{\rm 3}Key Laboratory of Intereactive Technology and Experience System, Ministry of Culture and Tourism, Beijing, China\\

    chenzn@bupt.edu.cn, $\{$luoxingshuang,wangweiqiu,zhaozc,sufei,menad$\}$@bupt.edu.cn
}

\begin{document}

\maketitle

\begin{abstract}
Recent Anomaly Detection (AD) methods have achieved great success with In-Distribution (ID) data. However, real-world data often exhibits distribution shift, causing huge performance decay on traditional AD methods. From this perspective, few previous work has explored AD with distribution shift, and the distribution-invariant normality learning has been proposed based on the Reverse Distillation (RD) framework. However, we observe the misalignment issue between the teacher and the student network that causes detection failure, thereby propose FiCo, \textbf{Fi}lter or \textbf{Co}mpensate, to address the distribution shift issue in AD. FiCo firstly compensates the distribution-specific information to reduce the misalignment between the teacher and student network via the Distribution-Specific Compensation (DiSCo) module, and secondly filters all abnormal information to capture distribution-invariant normality with the Distribution-Invariant Filter (DiIFi) module. Extensive experiments on three different AD benchmarks demonstrate the effectiveness of FiCo, which outperforms all existing state-of-the-art (SOTA) methods, and even achieves better results on the ID scenario compared with RD-based methods. Our code is available at https://github.com/znchen666/FiCo.
\end{abstract}

\section{Introduction}
Anomaly detection (AD) has been extensively researched and plays a critical role in numerous applications. Its main objective is to identify anomalous patterns within large amounts of data. Real-world applications, such as manufacturing quality control \cite{bergmann2019mvtec}, video surveillance \cite{liu2018future}, and medical monitoring \cite{schlegl2019f}, are in high demand for accurate and robust AD algorithms. In most scenarios, acquiring labeled anomaly data is challenging and expensive. As a result, unsupervised anomaly detection has become the prevailing focus of research. To address this issue, previous studies have made efforts from various aspects, such as reconstruction-based \cite{ristea2022self,zavrtanik2022dsr,zhang2023unsupervised,zhang2024realnet}, embedding-based \cite{roth2022towards,yu2021fastflow,huang2022registration,liu2023simplenet,zhu2024toward,lee2024text}, and knowledge distillation-based \cite{bergmann2020uninformed,salehi2021multiresolution,deng2022anomaly,tien2023revisiting,gu2023remembering} approaches, etc., which have led to significant advancements recently.

\begin{figure}
\begin{center}
\setlength{\belowcaptionskip}{-2pt}
\includegraphics[width=1.0\linewidth]{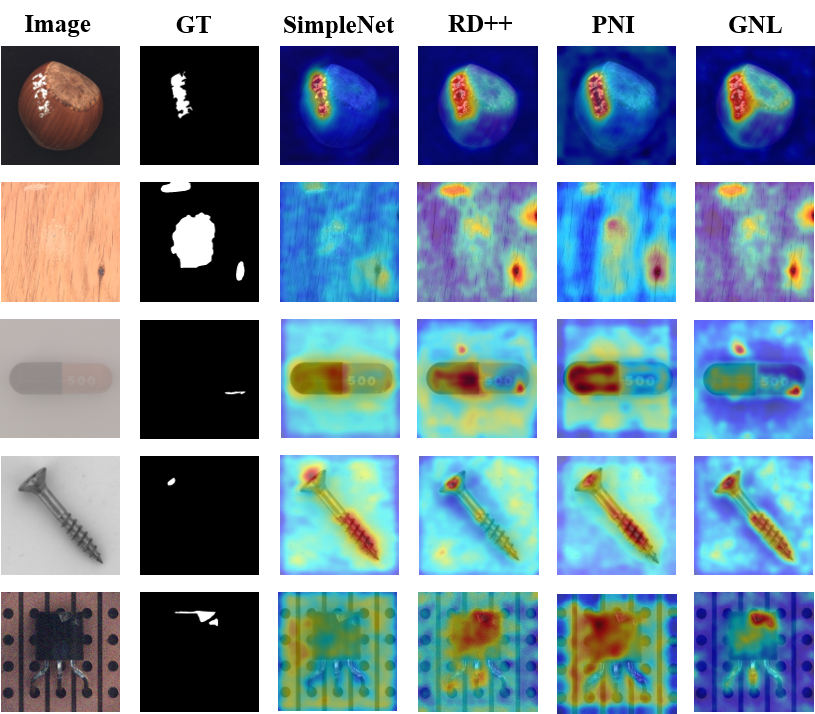}
    \caption{Anomaly map from different scenarios of SOTA AD methods \cite{bae2023pni,liu2023simplenet,tien2023revisiting,cao2023anomaly} on the MVTec benchmark \cite{bergmann2019mvtec}. The image of each row represents a different scenario, including ID and four OOD (brightness, contrast, defocus blur and gaussian noise) scenarios.}
  \label{fig:others}
\end{center}
\end{figure} 

These methods mostly assume training and test sets are In-Distribution (ID), thus the only purpose is to identify anomalies from normal ones without distribution shift. However, the assumption is not realistic in real-world scenarios, causing huge performance decay when confronting real-world Out-of-Distribution (OOD) data \cite{bae2023pni,liu2023simplenet,tien2023revisiting}, as shown in Fig. \ref{fig:others}. The test data possibly have both anomalous patterns and distribution shifts, and most methods show accurate anomaly map merely on the ID scenario without distribution shift, but are negatively affected by the distribution shift from OOD scenarios. From this perspective, we investigate the research for resolving the distribution shift between ID and OOD data on various downstream applications, such as image classification \cite{zhou2021domain,xu2021fourier,cha2021swad,mahajan2021domain,lv2022causality,chen2024practicaldg}, semantic segmentation \cite{choi2021robustnet,zhao2022style,huang2023style} and person re-identification \cite{liao2020interpretable,liao2022graph,chen2023cluster}, etc. The methods can be mainly categorized into data augmentation \cite{xu2021fourier,zhao2022style}, domain-invariant learning \cite{mahajan2021domain,lv2022causality}, and learning strategies \cite{cha2021swad,liao2022graph}, where domain-invariant learning has been the mainstream and achieves competitive results with relatively low computational costs.

Then \cite{cao2023anomaly} proposes the first work on AD under multiple OOD scenarios by designing the distribution-invariant normality learning. The method is based on the Reverse Distillation (RD) \cite{deng2022anomaly} framework, where the output from the teacher network flows to the student network via a one-class bottleneck module (OCBE). It learns invariant representation via consistency between multiple augmentations to filter distribution-specific information. However, we observe that there's information misalignment between the teacher and student network. Thus, the question emerges that ``What information should be filtered or compensated to obtain the invariant representation in the RD framework?"

This paper revisits the efficient and effective RD framework from the perspective of invariant representation to tackle the distribution shift issue. We observe that one drawback in \cite{cao2023anomaly} is the information loss on distribution-specific representation, while another drawback lies in the absence of an explicit mechanism to better learn distribution-invariant normality. Therefore, we firstly compensate for the distribution-specific information in the student network, while secondly filtering irrelevant information to achieve  distribution-invariant normality. In conclusion, this paper proposes Filter or Compensate (FiCo) method to thoroughly explore the invariant representation to resolve the distribution shift, surpassing state-of-the-art (SOTA) methods on multiple AD benchmarks with a relatively large margin.

The main contributions of the paper can be summarized as follows,
\begin{itemize}
    \item We propose FiCo for better invariant representation learning to address the distribution shift issue in AD task. Firstly, the Distribution-Invariant Filter (DiIFi) module is proposed to filter all abnormal information for distribution-invariant normality, including anomlous patterns and distribution-specific information.
    \item Secondly, the Distribution-Specific Compensation (DiSCo) module is designed to compensate for distribution-specific information, thereby reducing the misalignment between the teacher and student network. Consequently, merely the anomalous pattern is taken into account during the inference without the interference of the distribution-specific information. 
    \item Extensive experiments on different AD benchmarks with OOD scenarios manifest the superior performance of our method compared with SOTA AD methods. FiCo not only achieves better performance on OOD scenarios, but also improves accuracy on ID scenario to surpass RD-based methods. 
\end{itemize}

\section{Related Work}
\subsection{Anomaly Detection}

Recent anomaly detection methods have been dispersed into various categories, while the mainstream research can be coarsely categorized into reconstruction-based, embedding-based and knowledge distillation-based methods. For the reconstruction-based methods, AutoEncoder (AE) \cite{kingma2013auto} and Generative Adversarial Network (GAN) \cite{goodfellow2020generative} are widely adopted as the generative models to reconstruct samples. Then, research has been expanded to various aspects to address the issue, including the memory module \cite{park2020learning,hou2021divide,gu2023remembering}, pseudo-anomaly augmentation \cite{li2021cutpaste,schluter2022natural} and diffusion model \cite{wyatt2022anoddpm,zhang2023unsupervised,zhang2024realnet}, etc. Embedding-based methods show strong improvement in recent literatures by simply using pretrained networks for feature extraction. These methods identify anomalies by the input feature embedding with the normal feature distribution via different standards \cite{cohen2020sub,yu2021fastflow}. Meanwhile, different spatial feature are specifically designed for measurement \cite{defard2021padim,roth2022towards,bae2023pni,yao2023focus}. \cite{reiss2021panda,deecke2021transfer} introduce different modules for adaptation to the distribution of target dataset. Recently, text-based AD has emerged to leverage the capability of CLIP for textual knowledge with text prompts \cite{zhu2024toward,lee2024text}. 

Knowledge distillation is a promising solution for anomaly detection that the student network learns the anomaly-free feature from the teacher network and detects abnormal one based on the discrepancy. \cite{bergmann2020uninformed} ensembles multiple student networks for more discriminate feature. \cite{salehi2021multiresolution} designs feature-level distillation at various layers of the pretrained expert network. \cite{deng2022anomaly} proposes the reverse distillation framework that the one-class embedding from the teacher network flows to the student network to restore multi-scale feature. \cite{tien2023revisiting} improves \cite{deng2022anomaly} on feature compactness by designing optimal transport loss, and anomalous signal suppression by simulating pseudo-anomaly samples. \cite{gu2023remembering} designs the normality recall memory to store normal information based on RD framework to tackle ``normality forgetting" issue. \cite{cao2023anomaly} proposes distribution-invariant normality learning to tackle the distribution shift issue by introducing consistency loss on different augmented views.

\subsection{Out-of-Distribution Generalization}
Out-of-Distribution (OOD) issue is essential in various downstream tasks, where methods can be roughly categorized into three aspects, including data augmentation \cite{zhou2021domain,xu2021fourier}, domain-invariant learning \cite{mahajan2021domain,lv2022causality}, and learning strategies \cite{cha2021swad,liao2022graph}. Data augmentation has been a simple yet effective technique in OOD generalization by synthesizing novel images and features. Domain-invariant learning has been the mainstream solution for OOD generalization \cite{sun2016deep,lv2022causality}. Learning strategies such as meta learning \cite{li2018learning}, adversarial learning \cite{li2018domain}, gradient optimization \cite{foret2020sharpness} also boost the research on fundamental training protocols.

However, the AD task differs from those downstream tasks that no class or domain label is available, which cannot meet the requirements with many aforementioned solutions. From this perspective, domain-invariant learning is plausible and promising for its simplicity and effectiveness. Thus, \cite{cao2023anomaly} proposes the distribution-invariant normality learning by introducing common augmentations to filter distribution-specific information based on the RD \cite{deng2022anomaly} framework. Nevertheless, it neglects the mechanism of the RD framework with coarse consistency at different spatial levels, which harms the invariant representation from the student network.

\section{Preliminaries}
\subsection{Task Description}
Let $(x_{s}, y_{s})$, $(x_{t}, y_{t})$ denote the training and test samples with the label indicating anomalies, and suppose $\mathcal{X}_{id}$, $\mathcal{X}_{ood}$ are ID and OOD distributions. During the training process, the dataset merely contains normal samples with ID distribution, $\mathcal{I}_{s} = \left\{(x_{s}\in\mathcal{X}_{id} \mid y_{s}=0)\right\}$. However, in the inference stage, the test dataset contains both normal and anomalous samples with different distribution, $\mathcal{I}_{t} = \left\{(x_{t}\in\mathcal{X}_{id}\cup\mathcal{X}_{ood} \mid y_{t}={0,1})\right\}$, where ${0, 1}$ denote normal and anomalous samples, respectively. The goal of the task is to train models on the training datasets with only normal and ID samples, and generalize well on the unseen test dataset with anomalies and distribution shifts. Note that no data from test dataset is available during the training process.

\subsection{Briefly Review on RD-based Methods}
Reverse distillation for anomaly detection is first proposed in \cite{deng2022anomaly}, which consists of a frozen pretrained teacher network $E(\cdot)$, a one-class embedding (OCBE) module $\phi(\cdot)$ and a student network $D(\cdot)$. Unlike previous methods on knowledge distillation, the output of the pretrained teacher network is sequentially passed through the OCBE module and the student network, which means the high-level semantic knowledge flows to the student first. During the training process, cosine similarity is adopted to formulate the loss function,

\begin{equation}
    \mathcal{L}_{RD} = \sum_{k=1}^{K} \{1 - \frac{f^{E_k}\cdot f^{D_k}}{\Vert f^{E_k}\Vert \Vert f^{D_k}\Vert}\}
\label{base}
\end{equation}
where $K$ is the total number of layers, $E_{k}$ and $D_{k}$ are the $k^{th}$ layer of the encoder and the decoder, and $f^{E_k}$, $f^{D_k}$ are the feature maps of $x_{s}$ from the $k^{th}$ block of the teacher and student network, respectively. During inference, the multi-scale similarities of representation are utilized for evaluation, where low similarity score indicates anomalies.

Afterwards, GNL \cite{cao2023anomaly} proposes the distribution-invariant normality learning to tackle the distribution shift issue. It introduces multiple augmentations on the training sample to synthesize $x_{s}^{n}$, where $n$ is the $n^{th}$ augmented view of sample $x_{s}$. Then it designs consistency loss at both the OCBE module and the final output of the student network as $\mathcal{L}_{abs}$ and $\mathcal{L}_{lowf}$ to filter distribution-specific information,

\begin{equation}
    \mathcal{L}_{abs} = \sum_{n=1}^{N} \{1 - \frac{(f^{\phi})^{T}\cdot f^{\phi}_n}{\Vert f^{\phi}\Vert \Vert f^{\phi}_n\Vert}\}
\end{equation}
\begin{equation}
    \mathcal{L}_{lowf} = \sum_{n=1}^{N} \{1 - \frac{(f^{D_1})^{T}\cdot (f^{D_1}_n)}{\Vert f^{D_1}\Vert \Vert f^{D_1}_n\Vert}\}
\end{equation}
where $f^{\phi}$, $f^{\phi}_n$ are the output representation of original image and the $n^{th}$ augmented image from the OCBE module, while $f^{D}$, $f^{D}_n$ are the final output from the student network. Then the final loss can be formulated as,
\begin{equation}
    \mathcal{L}_{GNL} = \mathcal{L}_{RD} + \mathcal{L}_{abs} + \mathcal{L}_{lowf}
\end{equation}
During inference, \cite{cao2023anomaly} utilizes the existing Test-Time Augmentation (TTA) technique EFDM \cite{zhang2022exact} to minimize the discrepancy between the distribution of the test sample and the normal sample.

\section{Approach}
\begin{figure}
\begin{center}
  \setlength{\belowcaptionskip}{-2pt}
  \includegraphics[width=1\linewidth,height=0.25\textheight]{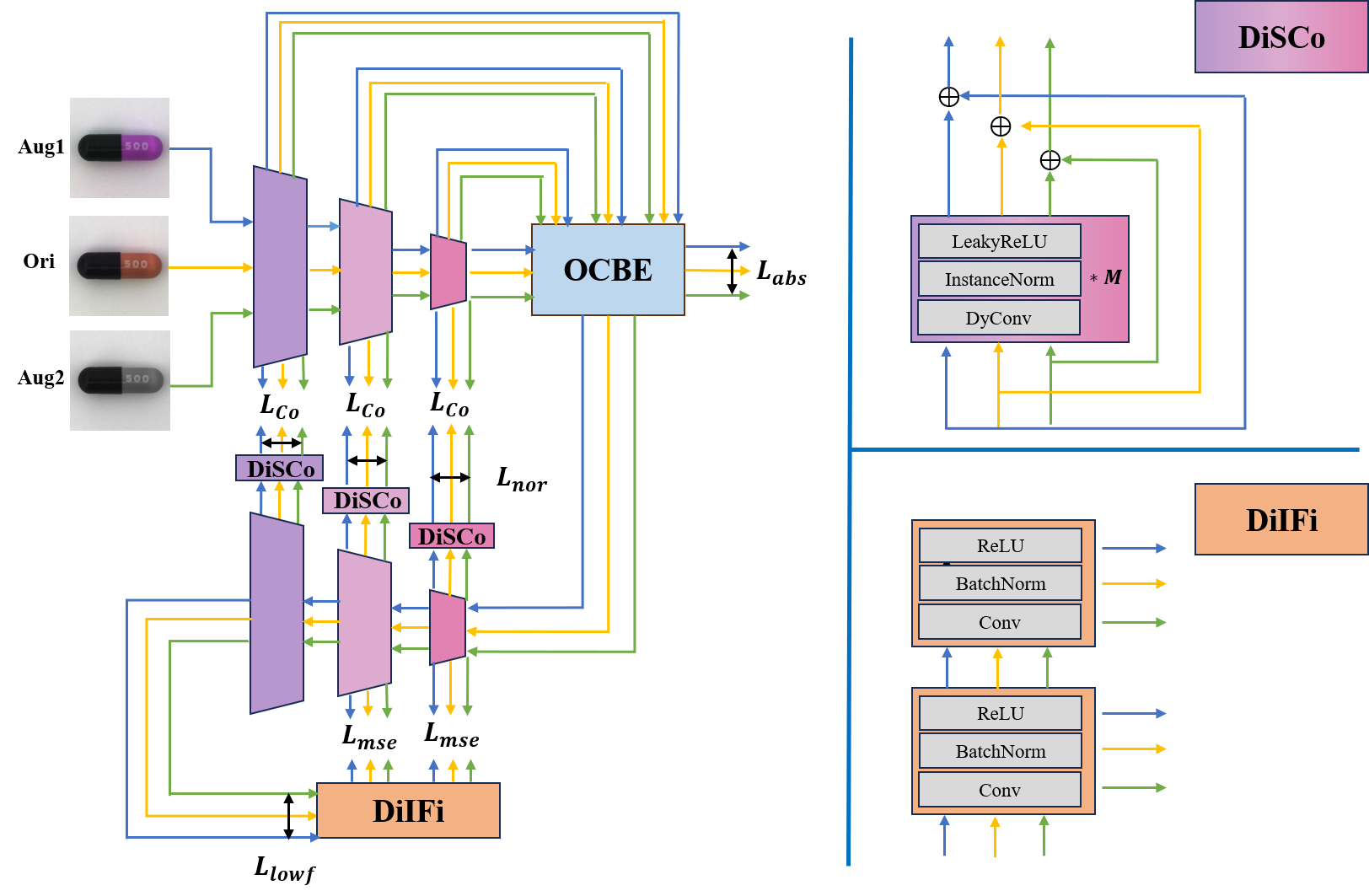}
    \caption{The overall architecture of our method FiCo, including detailed structure of DiSCo module and DiIFi module with designed losses. DiSCo module aims to compensate for the distribution-specific information by $\mathcal{L}_{Co}$ to prevent misalignment. DiIFi module attempts to filter abnormal patterns to obtain invariant normality with $\mathcal{L}_{Fi}$, including $\mathcal{L}_{lowf},\mathcal{L}_{mse},\mathcal{L}_{nor}$. $\mathcal{L}_{abs}$ indicates the consistency loss between original and augmented representation at OCBE module.}
  \label{fig:overall}
\end{center}
\end{figure}

Our method FiCo resolves the drawbacks of distribution-invariant normality learning in \cite{cao2023anomaly} by designing additional modules and loss functions. The overall architecture is presented in Fig. \ref{fig:overall}.

\subsection{Distribution-Specific Compensation Module}

GNL \cite{cao2023anomaly} designs training objectives to align the teacher and student network for distribution-invariant normality, but neglects the distribution shifts from the OOD samples in the test dataset. As a result, the misalignment of distribution-specific information confuses the model to misclassify the distribution shifts as anomalies. Therefore, we aim to compensate for the distribution-specific information to prevent the misleading effect of the distribution shifts. Specifically, assume that the representation map from each block of the student network consists of distribution-invariant information and distribution-specific information. As the inference process of RD framework is to calculate the multi-scale similarities between the representations of the pre-trained teacher network and the student network, the misalignment on distribution-specific information causes the discrepancy which is prone to be recognized as anomalous patterns. Under this observation, we investigate how to compensate for distribution-specific information to guarantee the alignment between the teacher and student network when inferring on OOD samples. Therefore, we propose the DiSCo module to take responsibility for preventing distribution-specific information loss. 

Suppose the DiSCo module as $C_k(\cdot)$ which is inserted after each block of the student network. The DiSCo module receives the feature map from each block and reconstructs distribution-specific information. To prevent representation loss, the shortcut is adopted and the final output can be described as,
\begin{equation}
    f_{F}^{D_k} = C_k(f^{D_k}) + f^{D_k}, \quad \quad f_{F,n}^{D_k} = C_k(f_n^{D_k}) + f_n^{D_k}
\end{equation}
where $f^{D_k}_{n}$ denotes the output of the $n^{th}$ augmented sample from the $k^{th}$ block of the student network, and $f_{F}^{D_k}$ and $f_{F,n}^{D_k}$ are the output from original and augmented views after the shortcut on DiSCo modules. Note that if $n$ is omitted, the symbol represents the original image without augmentation. Then we design the loss function to ensure the valid compensation, which can be formulated as,
\begin{equation}
    \mathcal{L}_{Co} = \sum_{k=1}^{K} \{1 - \frac{f^{E_k}\cdot f^{D_k}_{F}}{\Vert f^{E_k}\Vert \Vert f^{D_k}_{F}\Vert}\} + \alpha \sum_{n=1}^{N} \sum_{k=1}^{K} \{1 - \frac{f^{E_k}_{n}\cdot f^{D_k}_{F,n}}{\Vert f^{E_k}_{n}\Vert \Vert f^{D_k}_{F,n}\Vert}\}
\end{equation}
where $f^{E_k}_{n}$ is the output of the $n^{th}$ augmented sample from the $k^{th}$ block of the teacher network and $\alpha$ is the hyper-parameter to balance the ratio.

Furthermore, for the constitution of the DiSCo module, considering the discrepancy between the training dataset and the OOD test dataset, DyConv \cite{chen2020dynamic} is utilized to introduce attention to differentiate diverse feature distribution, followed with InstanceNorm \cite{ulyanov2016instance} and LeakyReLU \cite{xu2015empirical} for normalization and activation. The DiSCo module consists of $M$ blocks of DyConv, InstanceNorm, LeakyReLU and can be trained end-to-end in the student network.

\subsection{Distribution-Invariant Filter Module}
RD framework and the following improvement have made assumptions that anomalous patterns should be constrained towards the student network, so that the student network merely reconstructs normal patterns \cite{deng2022anomaly,tien2023revisiting}. As a result, the discrepancy between the teacher and student network can be maximized when confronting anomalies. From this perspective, we perceive that filtering all abnormal information in the student network to obtain the invariant normality naturally promotes invariant representation learning. However, GNL \cite{cao2023anomaly} designs $L_{lowf}$ at the final block of the student network for consistency on diverse augmented views of a single sample, but no explicit mechanisms on the previous blocks are exploited. Therefore, we attempt to incorporate the filter module at previous blocks of the student network to acquire invariant representation.

We analyze that the output of the final block is low-level information, including edges, colors, shapes, etc., where the consistency between different augmented views successfully captures invariant representation. However, the output from previous blocks are high-level semantics with distribution-specific information. As a result, directly applying the same function as $\mathcal{L}_{lowf}$ on previous blocks suffers from semantic information loss. Therefore, the distribution-invariant filter (DiIFi) module is designed to filter distribution-specific information from previous blocks, by imitating what the final DiSCo module $C_1(\cdot)$ recognizes as distribution-specific information. Specifically, let the DiIFi module as $I(\cdot)$ which consists of K-1 ConvBlocks (Convolution, BatchNorm, ReLU), and $I_k(\cdot)$ indicates the operation on the $k^{th}$ block to transform the distribution-specific information $ C_1(f_n^{D_1}) \in \mathbb{R}^{C\times H\times W}$ learned from the final DiSCo module. The formulation of sequential transformation of distribution-specific information to previous blocks is, 
\begin{equation}
f_{k,n}^{D_1} = \left\{
\begin{aligned}
& I_k(C_1(f_n^{D_1})) \in \mathbb{R}^{2C\times \frac{H}{2}\times \frac{W}{2}}, \quad & k = 2 \\
& I_k(f_{k-1,n}^{D_1}) \in \mathbb{R}^{2^{k-1}C\times \frac{H}{2^{k-1}}\times \frac{W}{2^{k-1}}}, \quad & k > 2 \\
\end{aligned}
\right.
\end{equation}
where $f_{k,n}^{D_1}$ denotes the transformed distribution-specific information of the $n^{th}$ augmented sample from the $k^{th}$ block of the student network.
The DiIFi module attempts to align the transformed distribution-specific feature $f_{k,n}^{D_1}$ with the corresponding compensated feature $C_k(f_n^{D_k})$ from previous DiSCo modules. Therefore, the Mean Square Error (MSE) loss is adopted to minimize the discrepancy,
\begin{equation}
    \mathcal{L}_{mse} = \sum_{k=2}^K {(C_k(f^{D_k}) - f_{k}^{D_1})}^2 + \sum_{n=1}^N \sum_{k=2}^K {(C_k(f_n^{D_k}) - f_{k,n}^{D_1})}^2
\end{equation}
where the first and the second item are operations on original and augmented views, respectively. The DiIFi module trained with $\mathcal{L}_{mse}$ has two-fold merits. Firstly, DiIFi module impels all previous DiSCo modules to mimic what the final DiSCo module learns. As the final representation consists of affluent low-level information, the transformation from which can prevent previous DiSCo modules to learn biased distribution-specific information. 
Secondly, as the DiSCo modules are supervised by $\mathcal{L}_{Co}$ to compensate for distribution-specific information, the DiIFi module implicitly promotes residual blocks to filter all abnormal information, including distribution-specific and anomalous patterns to learn invariant normality, as shown in Fig. \ref{fig:process}.

Lastly, to prevent the input of the DiIFi module from normality collapsing that the compensated distribution-specific information is biased from augmented representation, we also incorporate the consistency loss based on cosine similarity after the final DiSCo module, 
\begin{equation}
    \mathcal{L}_{nor} = \sum_{n=1}^{N} \{1 - \frac{(C_1(f^{D_1}))^{T}\cdot (C_1(f^{D_1}_n))}{\Vert C_1(f^{D_1})\Vert \Vert C_1(f^{D_1}_n)\Vert}\}
\end{equation}
The overall filter loss can be described as,
\begin{equation}
    \mathcal{L}_{Fi} = \mathcal{L}_{lowf} + \beta \mathcal{L}_{mse} + \gamma \mathcal{L}_{nor}
\end{equation}
where $\beta, \gamma$ are the balancing hyper-parameters.

\subsection{Training and Inference}
\textbf{Training.} The whole network is trained end-to-end with $L_{FiCo}$,
\begin{equation}    \mathcal{L}_{FiCo} = \mathcal{L}_{Fi} + \mathcal{L}_{abs} + \mathcal{L}_{Co}
\end{equation}
where we maintain the original framework from \cite{cao2023anomaly} to insert additional modules, and simply replace the $L_{RD}, L_{lowf}$ with $\mathcal{L}_{Fi}, \mathcal{L}_{Co}$ for the filter and compensation process. The overall algorithm is presented in the appendix.

\textbf{Inference.} During the inference process, all other settings are remained identical with \cite{cao2023anomaly}, including the test-time augmentation EFDM \cite{zhang2022exact} and the calculation process of sample-level anomaly score. The only difference is the additional DiSCo modules after all $K$ blocks from the student network are remained to compensate for the distribution-specific information, while the DiIFi module is discarded. 

\section{Experiments}
\subsection{Benchmarks}
Experiments are conducted on three AD benchmarks with distribution shifts \cite{cao2023anomaly}, including MVTec \cite{bergmann2019mvtec}, PACS \cite{li2017deeper} and CIFAR-10 \cite{krizhevsky2009learning}. MVTec is a widely-used industrial AD dataset with 15 categories, including 5 categories for texture anomalies and 10 categories for object anomalies. It consists of 5,354 images, including 3,629 normal images from the training set and 1,725 images from the test set with both normal and abnormal one. 
PACS is a prevalent dataset from OOD classification with a total of 9,991 images from seven classes and four domains. CIFAR-10 is used as the benchmark of the one-class novelty detection task, including 10 categories with 50,000 and 10,000 images from the training and test set. All datasets follow the same procedure in \cite{cao2023anomaly}. For MVTec and CIFAR-10, diverse visual corruptions are conducted to generate the OOD scenarios. For PACS, we merely use images on the common photo domain as the training set, and infer on different test sets from all domains.

\subsection{Implementation Details}
The backbone is WideResNet50 \cite{zagoruyko2016wide} as widely adopted and all the images for MVTec, PACS are resized to 256 $\times$ 256, while 32 $\times$ 32 for CIFAR-10. All the additional modules can be trained end-to-end with Adam optimizer \cite{kingma2014adam}, and the initial learning rate is set to 0.005. The hyper-parameters $\alpha, \beta, \gamma$ that control the balancing ratio of different additional losses are 0.05, 0.02, 1 for MVTec and CIFAR-10 dataset, while $\beta$ is set to 0.1 for PACS. The number of blocks $M$ in DiSCo module is set to 4 for all datasets. Other relevant hyper-parameters, such as the number of augmentations $N$, the style blending ratio in EFDM \cite{zhang2022exact}, and detailed operations are all maintained the same as \cite{cao2023anomaly} for fair comparison. 
For the evaluation metrics, the Area Under the Receiver Operator Curve (AUROC) on the sample-level is adopted \cite{cao2023anomaly}, which is a universal assessment between the normal and anomalous samples.

\subsection{Comparison with State-of-the-Art Methods}
We compare our proposed method FiCo with recent methods on anomaly detection, including Deep SVDD \cite{ruff2018deep}, f-AnoGAN \cite{schlegl2019f}, KD \cite{salehi2021multiresolution}, RD \cite{deng2022anomaly}, PatchCore \cite{roth2022towards}, RD++ \cite{tien2023revisiting}, SimpleNet \cite{liu2023simplenet}, PNI \cite{bae2023pni}, GNL \cite{cao2023anomaly}, RealNet \cite{zhang2024realnet}. The results are the average performance on all classes and the detailed results on each class are reported in the appendix. Note that $^\dag$ denotes our implementation and otherwise is the reported performance in \cite{cao2023anomaly}.

\textbf{MVTec.}
As shown in Table \ref{tab:mvtec}, our method FiCo surpasses all the prevalent methods on the average AUROC. Compared with SOTA methods PNI and RealNet, FiCo merely shows a slight decrease of 0.84$\%$ and 0.87$\%$ on ID performance, but improves 19.40$\%$ and 7.33$\%$ on the average performance of all OOD scenarios. Moreover, compared with all the RD-based methods, FiCo not only shows superiority on most OOD scenarios, but also has improved the performance on ID scenario to achieve SOTA performance. Furthermore, FiCo exceeds GNL on ID scenario with 1.04$\%$ and all OOD scenarios with an average of 1.22$\%$ that demonstrates the effectiveness of our method. 

\begin{table}[ht]
\small
  \setlength{\abovecaptionskip}{-3pt}
\caption{Comparison of state-of-the-art methods on sample-level AUROC for MVTec. ``Ori,Br,Co,Bl,No" represents original, brightness, contrast, defocus blur and gaussian noise scenario. }
\label{tab:mvtec}
\begin{center}
\setlength{\tabcolsep}{0.8pt}
\begin{tabular}{c|c|cccc|c}
\hline
\multirow{2}{*}{Method} & ID & \multicolumn{4}{c|}{OOD} & \multirow{2}{*}{Avg.}\\
\cline{2-6}
& Ori & Br & Co & Bl & No & \\
\hline
Deep SVDD \cite{ruff2018deep} & 70.0 & 55.2 & 50.1 & 68.8 & 59.1 & 60.6\\
f-AnoGAN \cite{schlegl2019f} & 75.7 & 48.4 & 49.3 & 38.0 & 39.1 & 50.1\\
KD \cite{salehi2021multiresolution} & 85.5 & 83.8 & 64.0 & 84.2 & 82.0 & 79.9\\
PatchCore$^\dag$ \cite{roth2022towards}& 99.1 & 96.0 & 92.1 & 97.2 & 93.9 & 95.7 \\
SimpleNet$^\dag$ \cite{liu2023simplenet} & 99.4 & 90.6 & 71.7 & 91.6 & 76.1 & 85.9\\
PNI$^\dag$ \cite{bae2023pni} & 99.6 & 87.8 & 67.6 & 90.2 & 66.1 & 82.3\\
RealNet$^\dag$ \cite{zhang2024realnet} & \textbf{99.7} & 92.3 & 95.4 & 95.6 & 76.7 & 91.9\\
\hline
RD \cite{deng2022anomaly} & 98.6 & 96.5 & 94.1 & \textbf{98.9} & 90.1 & 95.7\\
RD++$^\dag$ \cite{tien2023revisiting} & 98.7 & 96.1 & 95.2 & 98.2 & 84.4 & 94.5\\
GNL \cite{cao2023anomaly} & 98.0 & 97.4 & 97.5 & 97.8 & 94.1 & 97.0\\
GNL$^\dag$ \cite{cao2023anomaly} & 97.7 & 97.2 & 96.5 & 97.0 & 93.7 & 96.4\\
\textbf{FiCo (ours)} & 98.8 & \textbf{97.9} & \textbf{97.9} & 98.5 & \textbf{95.0} & \textbf{97.6}\\
\hline
\end{tabular}
\end{center}
\end{table}

\textbf{PACS.}
Results in Table \ref{tab:pacs} show that our method FiCo has superior generalizablity when confronting real-world data with OOD scenarios. All the prevalent AD methods suffer from huge performance decay, especially on the sketch domain, such as 14.84$\%$ drop on SimpleNet compared with our method FiCo. Instead, FiCo is capable to resolve the distribution shift issue that improves performance on all scenarios compared with RD-based methods, with 8.14$\%$ and 4.03$\%$ average improvement on RD and RD++. Meanwhile, FiCo surpasses GNL on all scenarios with an average of 2.94$\%$ that demonstrate the significance of our additional modules and losses.
\begin{table}[!ht]
\small
\setlength{\abovecaptionskip}{-3pt}
\caption{Comparison of state-of-the-art methods on sample-level AUROC for PACS. ``P,A,C,S" represents photo, art painting, cartoon and sketch domain.}
\label{tab:pacs}
\begin{center}
\setlength{\tabcolsep}{3pt}
\begin{tabular}{c|c|ccc|c}
\hline
\multirow{2}{*}{Method} & ID & \multicolumn{3}{c|}{OOD} & \multirow{2}{*}{Avg.}\\
\cline{2-5}
& P & A & C & S & \\
\hline
Deep SVDD \cite{ruff2018deep}& 40.9 & 53.4 & 41.2 & 39.5 & 43.8 \\
f-AnoGAN \cite{schlegl2019f}& 61.3 & 50.2 & 52.4 & \textbf{63.8} & 56.9 \\
KD \cite{salehi2021multiresolution} & 88.2 & 62.9 & 62.6 & 51.4 & 66.3\\
PatchCore$^\dag$ \cite{roth2022towards}& 77.5 & 57.5 & 56.5 & 52.1 & 60.9 \\
SimpleNet$^\dag$ \cite{liu2023simplenet}& \textbf{91.6} & 62.3 & 54.8 & 47.5 & 64.1\\
\hline
RD \cite{deng2022anomaly}& 81.5 & 61.1 & 60.3 & 55.1 & 64.5\\
RD++$^\dag$ \cite{tien2023revisiting}& 86.9 & 61.7 & 65.2 & 60.6 & 68.6\\
GNL \cite{cao2023anomaly}& 87.7 & 65.6 & 68.0 & 62.4 & 70.9\\
GNL$^\dag$ \cite{cao2023anomaly}& 87.5 & 64.8 & 68.3 & 58.1 & 69.7\\
\textbf{FiCo (ours)} & 89.7 & \textbf{67.6} & \textbf{70.9} & 62.3 & \textbf{72.6}\\
\hline
\end{tabular}
\end{center}
\end{table}

\textbf{CIFAR-10.}
Table \ref{tab:cifar10} presents the conventional one-class novelty detection task on CIFAR-10. One-class novelty detection is another form of anomaly detection where merely one class is regarded as the normal class and all other classes are abnormal counterparts. Our method FiCo also surpasses all methods on average AUROC. Especially compared with RD++ and GNL, FiCo shows superiority on both ID and OOD scenarios.

\begin{table}[ht]
\small
  \setlength{\abovecaptionskip}{-3pt}
\caption{Comparison of state-of-the-art methods on sample-level AUROC for CIFAR-10. ``Ori,Br,Co,Bl,No" represents original, brightness, contrast, defocus blur and gaussian noise scenario.}
\label{tab:cifar10}
\begin{center}
\setlength{\tabcolsep}{1.5pt}
\begin{tabular}{c|c|cccc|c}
\hline
\multirow{2}{*}{Method} & ID & \multicolumn{4}{c|}{OOD} & \multirow{2}{*}{Avg.}\\
\cline{2-6}
& Ori & Br & Co & Bl & No & \\
\hline
Deep SVDD \cite{ruff2018deep} & 64.6 & 59.1 & 55.9 & 62.1 & 54.5 & 59.3\\
f-AnoGAN \cite{schlegl2019f}& 70.3 & 54.6 & 57.2 & 60.7 & 51.8 & 58.9\\
KD \cite{salehi2021multiresolution}& 84.2 & 75.9 & 64.4 & 63.5 & 56.9 & 69.0 \\
PatchCore$^\dag$ \cite{roth2022towards}& 80.6 & 72.9 & 63.0 & 57.7 & 55.5 & 65.9\\
\hline
RD \cite{deng2022anomaly}& \textbf{84.6} & 75.9 & 65.3 & \textbf{66.7} & 58.8 & 70.3\\
RD++$^\dag$ \cite{tien2023revisiting}& 80.3 & 75.9 & 66.9 & 60.3 & 63.3 & 69.3\\
GNL \cite{cao2023anomaly}& 82.3 & 77.9 & 66.1 & 64.0 & 61.5 & 70.4\\
GNL$^\dag$ \cite{cao2023anomaly}& 79.2 & 76.9 & 67.5 & 63.2 & \textbf{64.6} & 70.3\\
\textbf{FiCo (ours)} & 80.5 & \textbf{77.8} & \textbf{69.2} & 63.8 & 64.4 & \textbf{71.1}\\
\hline
\end{tabular}
\end{center}
\end{table}

\subsection{Ablation Studies}
\textbf{Performance of different components.}
As our method FiCo consists of different additional modules and losses, we conduct ablation study on each part step-by-step to investigate the effectiveness. As shown in Table \ref{tab:component}, we start from the re-implementation of GNL \cite{cao2023anomaly} as the baseline method, and sequentially add DiSCo module with $\mathcal{L}_{Co}$, DiIFi module with $\mathcal{L}_{mse}$, and $\mathcal{L}_{nor}$. Note that except for $\mathcal{L}_{nor}$, other losses are omitted within the module design for abbreviation in Table \ref{tab:component}. Results show that each part can positively improve performance with 0.45$\%$, 1.59$\%$ and 0.90$\%$ on average. Specifically, after inserting the DiIFi module upon the DiSCo module, performance on all scenarios improves significantly for better invariant representation. Moreover, $\mathcal{L}_{nor}$ is capable to improve performance by preventing the normality collapsing.

\begin{table}[ht]
\small
  \setlength{\abovecaptionskip}{-3pt}
\caption{Effectiveness of different components on the PACS benchmark.}

\label{tab:component}
\begin{center}
\setlength{\tabcolsep}{3pt}
\begin{tabular}{c|c|ccc|c}
\hline
\multirow{2}{*}{Method} & ID & \multicolumn{3}{c|}{OOD} & \multirow{2}{*}{Avg.}\\
\cline{2-5}
& P & A & C & S & \\
\hline
GNL \cite{cao2023anomaly} & 87.5 & 64.8 & 68.3 & 58.1 & 69.7\\
DiSCo & 88.2 & 64.2 & 69.0 & 59.2 & 70.1 \\
DiSCo + DiIFi & 89.5 & 65.5 & 70.5 & 61.6 & 71.7\\
\textbf{FiCo} & \textbf{89.7} & \textbf{67.6} & \textbf{70.9} & \textbf{62.3} & \textbf{72.6}\\
\hline
\end{tabular}
\end{center}
\end{table}

\begin{figure}[ht]
\setlength{\belowcaptionskip}{-3pt}
  \centering
  \subcaptionbox{$\alpha$}{\includegraphics[width=0.31\linewidth]{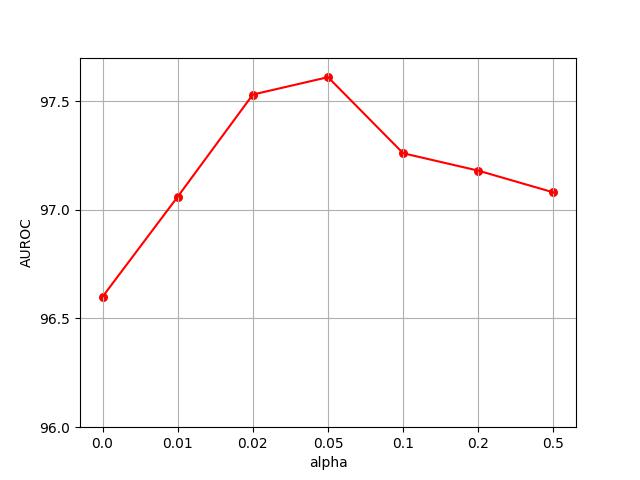}}
  \subcaptionbox{$\beta$}{\includegraphics[width=0.31\linewidth]{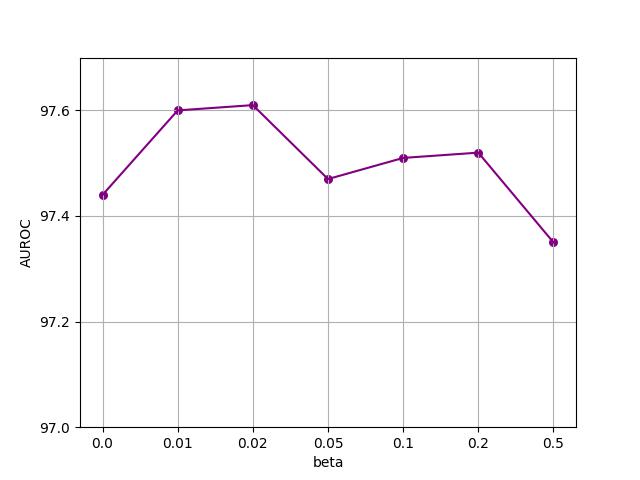}}
  \subcaptionbox{$\gamma$}{\includegraphics[width=0.31\linewidth]{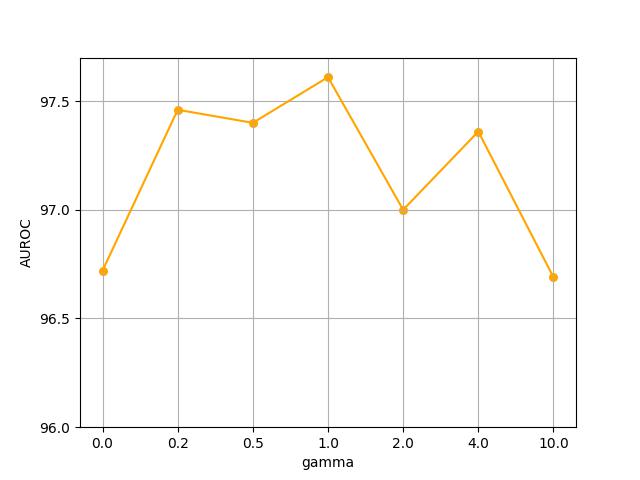}}
  \caption{Experimental results on hyper-parameters on the MVTec benchmark.}
  \label{fig:hyper}
\end{figure}

\textbf{Hyper-parameter Sensitivity.}
The three hyper-parameters $\alpha, \beta, \gamma$ are designed to balance the ratio between different additional losses. Ablation study is conducted to analyze the sensitivity to demonstrate the practicality of our method. Fig. \ref{fig:hyper} shows the results of each hyper-parameter with the others fixed to the optimal value. The fluctuations of three hyper-parameters are low with merely 0.55$\%$, 0.14$\%$ and 0.61$\%$, and any combination of hyper-parameter values can surpass GNL \cite{cao2023anomaly}. The best performance is achieved when $\alpha=0.05$, $\beta=0.02$ and $\gamma=1.0$.

\subsection{Analysis}
\begin{figure}[ht]
\begin{center}
\setlength{\belowcaptionskip}{-3pt}
  \includegraphics[width=1\linewidth,height=0.16\textheight]{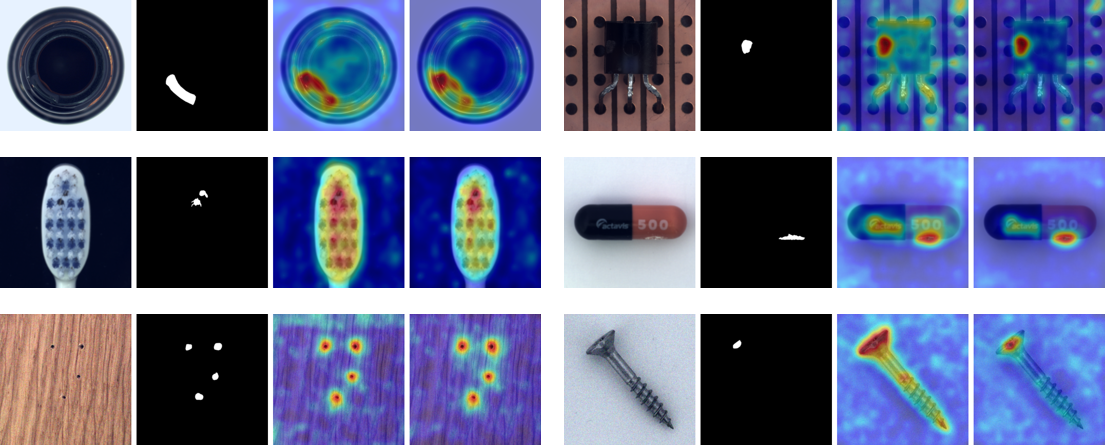}
    \caption{Anomaly map of $f^{D_k}$ and $f_F^{D_k}$ on the MVTec benchmark. Each row represents a different scenario, including ID, defocus blur and gaussian noise. For each scenario, two examples are shown with the original image, the groundtruth label, anomaly map from $f^{D_k}$ and from $f_F^{D_k}$.}
  \label{fig:process}
\end{center}
\end{figure}

\textbf{Filter and Compensation Process.}
Fig. \ref{fig:process} shows the anomaly map from the distribution-invariant representation $f^{D_k}$ for the filter process, and the final output $f_F^{D_k}$ for the compensation process. It can be clearly observed that the anomaly map from $f^{D_k}$ filters most of the abnormal information, including distribution-specific information and anomalous patterns. Thus, the discrepancy between $f^{D_k}$ and $f^{E_k}$ is maximized so that there exists more activated regions on both the real anomalous regions and distribution-specific regions, which validates the function of the DiIFi module to filter abnormal information. However, to prevent distribution-specific information to be recognized as anomalous patterns, the compensation process should restitute the distribution-specific noise, i.e., style information, for alignment with $f^{E_k}$. Compared with $f^{D_k}$, $f_F^{D_k}$ merely concentrates on the anomalous regions and the activated distribution-specific regions are mostly eliminated. Furthermore, our method not only resolves the distribution shift issue on OOD scenarios, but also shows strong robustness on ID data during the filter and compensation process. 

\begin{figure}[H]
\begin{center}
\setlength{\belowcaptionskip}{-3pt}
  \includegraphics[width=1\linewidth]{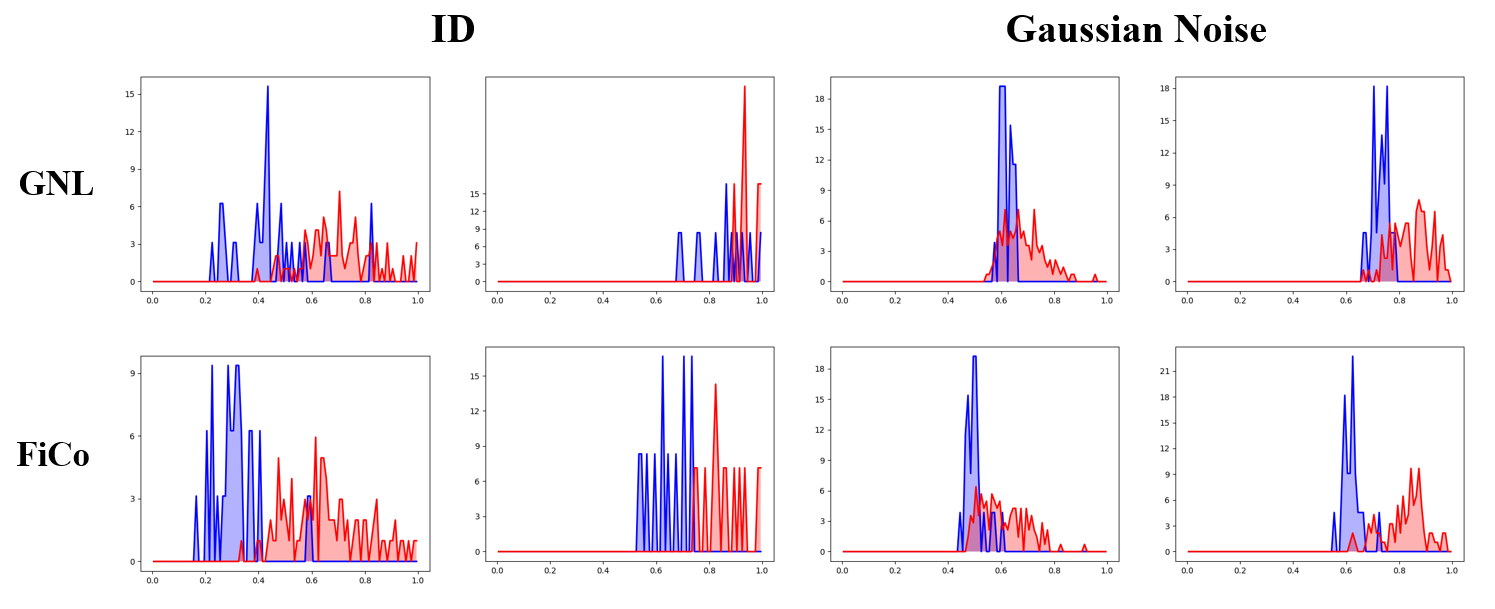}
    \caption{Anomaly scores of FiCo and GNL \cite{cao2023anomaly} on ID, gaussian noise scenario from the MVTec benchmark. Each scenario consists of two examples from 'zipper', 'toothbrush', and 'pill', 'metal nut'. Color blue and red indicates distribution on normal samples and anomalous samples, respectively.}
  \label{fig:score}
\end{center}
\end{figure}

\textbf{Key Difference for Anomaly Score.}
Anomaly detection aims to enlarge the discrepancy on the distribution of anomaly scores between normal samples and anomalies. Thus, we visualize the distribution of sample-level anomaly scores on all images on the hardest scenario gaussian noise and the simplest scenario ID. Fig. \ref{fig:score} displays the comparison between FiCo and GNL \cite{cao2023anomaly} to illustrate our merits. It can be observed that GNL fails on several cases that the overlap between the normal samples and anomalies is large. Instead, FiCo not only increases the discrepancy between normal samples and anomalies on OOD scenarios, but also shows strong adaptability on ID data. Furthermore, the anomaly scores for FiCo on normal samples are smaller than GNL that proves the effectiveness.

\textbf{Time Consumption.}
We evaluate the total training and inference time on SOTA RD-based methods. As presented in Table \ref{tab:time}, the total training time is approximately three times shorter than RD++ \cite{tien2023revisiting} for a single epoch, let alone RD++ uses 280 epochs while FiCo merely consumes 20 epochs. Meanwhile, our method consumes longer training time compared with GNL \cite{cao2023anomaly} due to the additional modules but remains reasonable. For the inference time, our method FiCo takes slightly longer time than RD++ and GNL, but yields substantial improvement on all ID and OOD scenarios to achieve SOTA performance. 

\begin{table}[htb]
\small
  \setlength{\abovecaptionskip}{-3pt}
\caption{Comparison of time consumption for RD-based methods on category 'toothbrush' in the MVTec benchmark on TESLA T4 GPU. $T_{train}$ and $T_{test}$ are training time per epoch and test time per image.}
\label{tab:time}
\begin{center}
\setlength{\tabcolsep}{3pt}
\begin{tabular}{l|c|c|c}
    \hline
    \multirow{2}{*}{Method} & {$T_{train}$} & {$T_{test}$} & {Results}\\
    & \multicolumn{1}{c|}{(s/epoch)} & (ms/image) & ($\%$)\\
  \hline
  RD++ \cite{tien2023revisiting} & 59.6 & 31.3 & 95.8\\
  GNL \cite{cao2023anomaly} & 13.3 & 40.8 & 95.2\\
  \textbf{FiCo (ours)} & 18.1 & 46.2 & 97.5\\
  \hline
\end{tabular}
\end{center}
\end{table}

\section{Conclusion}
This paper proposes FiCo to tackle the misalignment issue from the perspective of invariant representation for anomaly detection under distribution shifts. With the filter and compensation process, the model is capable to learn distribution-invariant normality and identify real anomalous patterns rather than distribution-specific information. The proposed DiSCo and DiIFi modules with novel training objectives are exclusively designed to address the misalignment issue, and can also be trained end-to-end without much extra costs. Experiments on several benchmarks for both industrial anomaly detection and one-class novelty detection in AD, show the strong generalizability and robustness of our method. Moreover, the visualizations are valid for the explanation of the whole process. In the future, the exploration of large vision-language models and generative models to resolve the distribution shift in anomaly detection worth more exploration.

\section{Acknowledgment}
\quad This work is supported by Chinese National Natural Science Foundation under Grants (62076033).

\bibliography{aaai25}

\begin{thebibliography}{57}
\providecommand{\natexlab}[1]{#1}

\bibitem[{Bae, Lee, and Kim(2023)}]{bae2023pni}
Bae, J.; Lee, J.-H.; and Kim, S. 2023.
\newblock Pni: industrial anomaly detection using position and neighborhood information.
\newblock In \emph{Proceedings of the IEEE/CVF International Conference on Computer Vision}, 6373--6383.

\bibitem[{Bergmann et~al.(2019)Bergmann, Fauser, Sattlegger, and Steger}]{bergmann2019mvtec}
Bergmann, P.; Fauser, M.; Sattlegger, D.; and Steger, C. 2019.
\newblock MVTec AD--A comprehensive real-world dataset for unsupervised anomaly detection.
\newblock In \emph{Proceedings of the IEEE/CVF conference on computer vision and pattern recognition}, 9592--9600.

\bibitem[{Bergmann et~al.(2020)Bergmann, Fauser, Sattlegger, and Steger}]{bergmann2020uninformed}
Bergmann, P.; Fauser, M.; Sattlegger, D.; and Steger, C. 2020.
\newblock Uninformed students: Student-teacher anomaly detection with discriminative latent embeddings.
\newblock In \emph{Proceedings of the IEEE/CVF conference on computer vision and pattern recognition}, 4183--4192.

\bibitem[{Cao, Zhu, and Pang(2023)}]{cao2023anomaly}
Cao, T.; Zhu, J.; and Pang, G. 2023.
\newblock Anomaly detection under distribution shift.
\newblock In \emph{Proceedings of the IEEE/CVF International Conference on Computer Vision}, 6511--6523.

\bibitem[{Cha et~al.(2021)Cha, Chun, Lee, Cho, Park, Lee, and Park}]{cha2021swad}
Cha, J.; Chun, S.; Lee, K.; Cho, H.-C.; Park, S.; Lee, Y.; and Park, S. 2021.
\newblock Swad: Domain generalization by seeking flat minima.
\newblock \emph{Advances in Neural Information Processing Systems}, 34: 22405--22418.

\bibitem[{Chen et~al.(2020)Chen, Dai, Liu, Chen, Yuan, and Liu}]{chen2020dynamic}
Chen, Y.; Dai, X.; Liu, M.; Chen, D.; Yuan, L.; and Liu, Z. 2020.
\newblock Dynamic convolution: Attention over convolution kernels.
\newblock In \emph{Proceedings of the IEEE/CVF conference on computer vision and pattern recognition}, 11030--11039.

\bibitem[{Chen et~al.(2023)Chen, Wang, Zhao, Su, Men, and Dong}]{chen2023cluster}
Chen, Z.; Wang, W.; Zhao, Z.; Su, F.; Men, A.; and Dong, Y. 2023.
\newblock Cluster-instance normalization: A statistical relation-aware normalization for generalizable person re-identification.
\newblock \emph{IEEE Transactions on Multimedia}.

\bibitem[{Chen et~al.(2024)Chen, Wang, Zhao, Su, Men, and Meng}]{chen2024practicaldg}
Chen, Z.; Wang, W.; Zhao, Z.; Su, F.; Men, A.; and Meng, H. 2024.
\newblock PracticalDG: Perturbation Distillation on Vision-Language Models for Hybrid Domain Generalization.
\newblock In \emph{Proceedings of the IEEE/CVF Conference on Computer Vision and Pattern Recognition}, 23501--23511.

\bibitem[{Choi et~al.(2021)Choi, Jung, Yun, Kim, Kim, and Choo}]{choi2021robustnet}
Choi, S.; Jung, S.; Yun, H.; Kim, J.~T.; Kim, S.; and Choo, J. 2021.
\newblock Robustnet: Improving domain generalization in urban-scene segmentation via instance selective whitening.
\newblock In \emph{Proceedings of the IEEE/CVF Conference on Computer Vision and Pattern Recognition}, 11580--11590.

\bibitem[{Cohen and Hoshen(2020)}]{cohen2020sub}
Cohen, N.; and Hoshen, Y. 2020.
\newblock Sub-image anomaly detection with deep pyramid correspondences.
\newblock \emph{arXiv preprint arXiv:2005.02357}.

\bibitem[{Deecke et~al.(2021)Deecke, Ruff, Vandermeulen, and Bilen}]{deecke2021transfer}
Deecke, L.; Ruff, L.; Vandermeulen, R.~A.; and Bilen, H. 2021.
\newblock Transfer-based semantic anomaly detection.
\newblock In \emph{International Conference on Machine Learning}, 2546--2558. PMLR.

\bibitem[{Defard et~al.(2021)Defard, Setkov, Loesch, and Audigier}]{defard2021padim}
Defard, T.; Setkov, A.; Loesch, A.; and Audigier, R. 2021.
\newblock Padim: a patch distribution modeling framework for anomaly detection and localization.
\newblock In \emph{International Conference on Pattern Recognition}, 475--489. Springer.

\bibitem[{Deng and Li(2022)}]{deng2022anomaly}
Deng, H.; and Li, X. 2022.
\newblock Anomaly detection via reverse distillation from one-class embedding.
\newblock In \emph{Proceedings of the IEEE/CVF Conference on Computer Vision and Pattern Recognition}, 9737--9746.

\bibitem[{Foret et~al.(2020)Foret, Kleiner, Mobahi, and Neyshabur}]{foret2020sharpness}
Foret, P.; Kleiner, A.; Mobahi, H.; and Neyshabur, B. 2020.
\newblock Sharpness-aware minimization for efficiently improving generalization.
\newblock \emph{arXiv preprint arXiv:2010.01412}.

\bibitem[{Goodfellow et~al.(2020)Goodfellow, Pouget-Abadie, Mirza, Xu, Warde-Farley, Ozair, Courville, and Bengio}]{goodfellow2020generative}
Goodfellow, I.; Pouget-Abadie, J.; Mirza, M.; Xu, B.; Warde-Farley, D.; Ozair, S.; Courville, A.; and Bengio, Y. 2020.
\newblock Generative adversarial networks.
\newblock \emph{Communications of the ACM}, 63(11): 139--144.

\bibitem[{Gu et~al.(2023)Gu, Liu, Chen, Yi, Zhang, Wang, Wang, Shu, Jiang, and Ma}]{gu2023remembering}
Gu, Z.; Liu, L.; Chen, X.; Yi, R.; Zhang, J.; Wang, Y.; Wang, C.; Shu, A.; Jiang, G.; and Ma, L. 2023.
\newblock Remembering Normality: Memory-guided Knowledge Distillation for Unsupervised Anomaly Detection.
\newblock In \emph{Proceedings of the IEEE/CVF International Conference on Computer Vision}, 16401--16409.

\bibitem[{Hou et~al.(2021)Hou, Zhang, Zhong, Xie, Pu, and Zhou}]{hou2021divide}
Hou, J.; Zhang, Y.; Zhong, Q.; Xie, D.; Pu, S.; and Zhou, H. 2021.
\newblock Divide-and-assemble: Learning block-wise memory for unsupervised anomaly detection.
\newblock In \emph{Proceedings of the IEEE/CVF International Conference on Computer Vision}, 8791--8800.

\bibitem[{Huang et~al.(2022)Huang, Guan, Jiang, Zhang, Spratling, and Wang}]{huang2022registration}
Huang, C.; Guan, H.; Jiang, A.; Zhang, Y.; Spratling, M.; and Wang, Y.-F. 2022.
\newblock Registration based few-shot anomaly detection.
\newblock In \emph{European Conference on Computer Vision}, 303--319. Springer.

\bibitem[{Huang et~al.(2023)Huang, Chen, Li, Li, Li, Song, Yan, and Xiong}]{huang2023style}
Huang, W.; Chen, C.; Li, Y.; Li, J.; Li, C.; Song, F.; Yan, Y.; and Xiong, Z. 2023.
\newblock Style Projected Clustering for Domain Generalized Semantic Segmentation.
\newblock In \emph{Proceedings of the IEEE/CVF Conference on Computer Vision and Pattern Recognition}, 3061--3071.

\bibitem[{Kingma and Ba(2014)}]{kingma2014adam}
Kingma, D.~P.; and Ba, J. 2014.
\newblock Adam: A method for stochastic optimization.
\newblock \emph{arXiv preprint arXiv:1412.6980}.

\bibitem[{Kingma and Welling(2013)}]{kingma2013auto}
Kingma, D.~P.; and Welling, M. 2013.
\newblock Auto-encoding variational bayes.
\newblock \emph{arXiv preprint arXiv:1312.6114}.

\bibitem[{Krizhevsky, Hinton et~al.(2009)}]{krizhevsky2009learning}
Krizhevsky, A.; Hinton, G.; et~al. 2009.
\newblock Learning multiple layers of features from tiny images.

\bibitem[{Lee and Choi(2024)}]{lee2024text}
Lee, M.; and Choi, J. 2024.
\newblock Text-Guided Variational Image Generation for Industrial Anomaly Detection and Segmentation.
\newblock In \emph{Proceedings of the IEEE/CVF Conference on Computer Vision and Pattern Recognition}, 26519--26528.

\bibitem[{Li et~al.(2021)Li, Sohn, Yoon, and Pfister}]{li2021cutpaste}
Li, C.-L.; Sohn, K.; Yoon, J.; and Pfister, T. 2021.
\newblock Cutpaste: Self-supervised learning for anomaly detection and localization.
\newblock In \emph{Proceedings of the IEEE/CVF conference on computer vision and pattern recognition}, 9664--9674.

\bibitem[{Li et~al.(2018{\natexlab{a}})Li, Yang, Song, and Hospedales}]{li2018learning}
Li, D.; Yang, Y.; Song, Y.-Z.; and Hospedales, T. 2018{\natexlab{a}}.
\newblock Learning to generalize: Meta-learning for domain generalization.
\newblock In \emph{Proceedings of the AAAI conference on artificial intelligence}, volume~32.

\bibitem[{Li et~al.(2017)Li, Yang, Song, and Hospedales}]{li2017deeper}
Li, D.; Yang, Y.; Song, Y.-Z.; and Hospedales, T.~M. 2017.
\newblock Deeper, broader and artier domain generalization.
\newblock In \emph{Proceedings of the IEEE international conference on computer vision}, 5542--5550.

\bibitem[{Li et~al.(2018{\natexlab{b}})Li, Pan, Wang, and Kot}]{li2018domain}
Li, H.; Pan, S.~J.; Wang, S.; and Kot, A.~C. 2018{\natexlab{b}}.
\newblock Domain generalization with adversarial feature learning.
\newblock In \emph{Proceedings of the IEEE conference on computer vision and pattern recognition}, 5400--5409.

\bibitem[{Liao and Shao(2020)}]{liao2020interpretable}
Liao, S.; and Shao, L. 2020.
\newblock Interpretable and generalizable person re-identification with query-adaptive convolution and temporal lifting.
\newblock In \emph{Computer Vision--ECCV 2020: 16th European Conference, Glasgow, UK, August 23--28, 2020, Proceedings, Part XI 16}, 456--474. Springer.

\bibitem[{Liao and Shao(2022)}]{liao2022graph}
Liao, S.; and Shao, L. 2022.
\newblock Graph sampling based deep metric learning for generalizable person re-identification.
\newblock In \emph{Proceedings of the IEEE/CVF Conference on Computer Vision and Pattern Recognition}, 7359--7368.

\bibitem[{Liu et~al.(2018)Liu, Luo, Lian, and Gao}]{liu2018future}
Liu, W.; Luo, W.; Lian, D.; and Gao, S. 2018.
\newblock Future frame prediction for anomaly detection--a new baseline.
\newblock In \emph{Proceedings of the IEEE conference on computer vision and pattern recognition}, 6536--6545.

\bibitem[{Liu et~al.(2023)Liu, Zhou, Xu, and Wang}]{liu2023simplenet}
Liu, Z.; Zhou, Y.; Xu, Y.; and Wang, Z. 2023.
\newblock Simplenet: A simple network for image anomaly detection and localization.
\newblock In \emph{Proceedings of the IEEE/CVF Conference on Computer Vision and Pattern Recognition}, 20402--20411.

\bibitem[{Lv et~al.(2022)Lv, Liang, Li, Zang, Liu, Wang, and Liu}]{lv2022causality}
Lv, F.; Liang, J.; Li, S.; Zang, B.; Liu, C.~H.; Wang, Z.; and Liu, D. 2022.
\newblock Causality inspired representation learning for domain generalization.
\newblock In \emph{Proceedings of the IEEE/CVF Conference on Computer Vision and Pattern Recognition}, 8046--8056.

\bibitem[{Mahajan, Tople, and Sharma(2021)}]{mahajan2021domain}
Mahajan, D.; Tople, S.; and Sharma, A. 2021.
\newblock Domain generalization using causal matching.
\newblock In \emph{International Conference on Machine Learning}, 7313--7324. PMLR.

\bibitem[{Park, Noh, and Ham(2020)}]{park2020learning}
Park, H.; Noh, J.; and Ham, B. 2020.
\newblock Learning memory-guided normality for anomaly detection.
\newblock In \emph{Proceedings of the IEEE/CVF conference on computer vision and pattern recognition}, 14372--14381.

\bibitem[{Reiss et~al.(2021)Reiss, Cohen, Bergman, and Hoshen}]{reiss2021panda}
Reiss, T.; Cohen, N.; Bergman, L.; and Hoshen, Y. 2021.
\newblock Panda: Adapting pretrained features for anomaly detection and segmentation.
\newblock In \emph{Proceedings of the IEEE/CVF Conference on Computer Vision and Pattern Recognition}, 2806--2814.

\bibitem[{Ristea et~al.(2022)Ristea, Madan, Ionescu, Nasrollahi, Khan, Moeslund, and Shah}]{ristea2022self}
Ristea, N.-C.; Madan, N.; Ionescu, R.~T.; Nasrollahi, K.; Khan, F.~S.; Moeslund, T.~B.; and Shah, M. 2022.
\newblock Self-supervised predictive convolutional attentive block for anomaly detection.
\newblock In \emph{Proceedings of the IEEE/CVF conference on computer vision and pattern recognition}, 13576--13586.

\bibitem[{Roth et~al.(2022)Roth, Pemula, Zepeda, Sch{\"o}lkopf, Brox, and Gehler}]{roth2022towards}
Roth, K.; Pemula, L.; Zepeda, J.; Sch{\"o}lkopf, B.; Brox, T.; and Gehler, P. 2022.
\newblock Towards total recall in industrial anomaly detection.
\newblock In \emph{Proceedings of the IEEE/CVF Conference on Computer Vision and Pattern Recognition}, 14318--14328.

\bibitem[{Ruff et~al.(2018)Ruff, Vandermeulen, Goernitz, Deecke, Siddiqui, Binder, M{\"u}ller, and Kloft}]{ruff2018deep}
Ruff, L.; Vandermeulen, R.; Goernitz, N.; Deecke, L.; Siddiqui, S.~A.; Binder, A.; M{\"u}ller, E.; and Kloft, M. 2018.
\newblock Deep one-class classification.
\newblock In \emph{International conference on machine learning}, 4393--4402. PMLR.

\bibitem[{Salehi et~al.(2021)Salehi, Sadjadi, Baselizadeh, Rohban, and Rabiee}]{salehi2021multiresolution}
Salehi, M.; Sadjadi, N.; Baselizadeh, S.; Rohban, M.~H.; and Rabiee, H.~R. 2021.
\newblock Multiresolution knowledge distillation for anomaly detection.
\newblock In \emph{Proceedings of the IEEE/CVF conference on computer vision and pattern recognition}, 14902--14912.

\bibitem[{Schlegl et~al.(2019)Schlegl, Seeb{\"o}ck, Waldstein, Langs, and Schmidt-Erfurth}]{schlegl2019f}
Schlegl, T.; Seeb{\"o}ck, P.; Waldstein, S.~M.; Langs, G.; and Schmidt-Erfurth, U. 2019.
\newblock f-AnoGAN: Fast unsupervised anomaly detection with generative adversarial networks.
\newblock \emph{Medical image analysis}, 54: 30--44.

\bibitem[{Schl{\"u}ter et~al.(2022)Schl{\"u}ter, Tan, Hou, and Kainz}]{schluter2022natural}
Schl{\"u}ter, H.~M.; Tan, J.; Hou, B.; and Kainz, B. 2022.
\newblock Natural synthetic anomalies for self-supervised anomaly detection and localization.
\newblock In \emph{European Conference on Computer Vision}, 474--489. Springer.

\bibitem[{Sun and Saenko(2016)}]{sun2016deep}
Sun, B.; and Saenko, K. 2016.
\newblock Deep coral: Correlation alignment for deep domain adaptation.
\newblock In \emph{Computer Vision--ECCV 2016 Workshops: Amsterdam, The Netherlands, October 8-10 and 15-16, 2016, Proceedings, Part III 14}, 443--450. Springer.

\bibitem[{Tien et~al.(2023)Tien, Nguyen, Tran, Huy, Duong, Nguyen, and Truong}]{tien2023revisiting}
Tien, T.~D.; Nguyen, A.~T.; Tran, N.~H.; Huy, T.~D.; Duong, S.; Nguyen, C. D.~T.; and Truong, S.~Q. 2023.
\newblock Revisiting reverse distillation for anomaly detection.
\newblock In \emph{Proceedings of the IEEE/CVF Conference on Computer Vision and Pattern Recognition}, 24511--24520.

\bibitem[{Ulyanov, Vedaldi, and Lempitsky(2016)}]{ulyanov2016instance}
Ulyanov, D.; Vedaldi, A.; and Lempitsky, V. 2016.
\newblock Instance normalization: The missing ingredient for fast stylization.
\newblock \emph{arXiv preprint arXiv:1607.08022}.

\bibitem[{Wyatt et~al.(2022)Wyatt, Leach, Schmon, and Willcocks}]{wyatt2022anoddpm}
Wyatt, J.; Leach, A.; Schmon, S.~M.; and Willcocks, C.~G. 2022.
\newblock Anoddpm: Anomaly detection with denoising diffusion probabilistic models using simplex noise.
\newblock In \emph{Proceedings of the IEEE/CVF Conference on Computer Vision and Pattern Recognition}, 650--656.

\bibitem[{Xu et~al.(2015)Xu, Wang, Chen, and Li}]{xu2015empirical}
Xu, B.; Wang, N.; Chen, T.; and Li, M. 2015.
\newblock Empirical evaluation of rectified activations in convolutional network.
\newblock \emph{arXiv preprint arXiv:1505.00853}.

\bibitem[{Xu et~al.(2021)Xu, Zhang, Zhang, Wang, and Tian}]{xu2021fourier}
Xu, Q.; Zhang, R.; Zhang, Y.; Wang, Y.; and Tian, Q. 2021.
\newblock A fourier-based framework for domain generalization.
\newblock In \emph{Proceedings of the IEEE/CVF Conference on Computer Vision and Pattern Recognition}, 14383--14392.

\bibitem[{Yao et~al.(2023)Yao, Li, Qian, Luo, and Zhang}]{yao2023focus}
Yao, X.; Li, R.; Qian, Z.; Luo, Y.; and Zhang, C. 2023.
\newblock Focus the discrepancy: Intra-and inter-correlation learning for image anomaly detection.
\newblock In \emph{Proceedings of the IEEE/CVF International Conference on Computer Vision}, 6803--6813.

\bibitem[{Yu et~al.(2021)Yu, Zheng, Wang, Li, Wu, Zhao, and Wu}]{yu2021fastflow}
Yu, J.; Zheng, Y.; Wang, X.; Li, W.; Wu, Y.; Zhao, R.; and Wu, L. 2021.
\newblock Fastflow: Unsupervised anomaly detection and localization via 2d normalizing flows.
\newblock \emph{arXiv preprint arXiv:2111.07677}.

\bibitem[{Zagoruyko and Komodakis(2016)}]{zagoruyko2016wide}
Zagoruyko, S.; and Komodakis, N. 2016.
\newblock Wide residual networks.
\newblock \emph{arXiv preprint arXiv:1605.07146}.

\bibitem[{Zavrtanik, Kristan, and Sko{\v{c}}aj(2022)}]{zavrtanik2022dsr}
Zavrtanik, V.; Kristan, M.; and Sko{\v{c}}aj, D. 2022.
\newblock Dsr--a dual subspace re-projection network for surface anomaly detection.
\newblock In \emph{European conference on computer vision}, 539--554. Springer.

\bibitem[{Zhang et~al.(2023)Zhang, Li, Li, Dai, Jiang, and Xia}]{zhang2023unsupervised}
Zhang, X.; Li, N.; Li, J.; Dai, T.; Jiang, Y.; and Xia, S.-T. 2023.
\newblock Unsupervised surface anomaly detection with diffusion probabilistic model.
\newblock In \emph{Proceedings of the IEEE/CVF International Conference on Computer Vision}, 6782--6791.

\bibitem[{Zhang, Xu, and Zhou(2024)}]{zhang2024realnet}
Zhang, X.; Xu, M.; and Zhou, X. 2024.
\newblock RealNet: A feature selection network with realistic synthetic anomaly for anomaly detection.
\newblock In \emph{Proceedings of the IEEE/CVF Conference on Computer Vision and Pattern Recognition}, 16699--16708.

\bibitem[{Zhang et~al.(2022)Zhang, Li, Li, Jia, and Zhang}]{zhang2022exact}
Zhang, Y.; Li, M.; Li, R.; Jia, K.; and Zhang, L. 2022.
\newblock Exact feature distribution matching for arbitrary style transfer and domain generalization.
\newblock In \emph{Proceedings of the IEEE/CVF Conference on Computer Vision and Pattern Recognition}, 8035--8045.

\bibitem[{Zhao et~al.(2022)Zhao, Zhong, Zhao, Sebe, and Lee}]{zhao2022style}
Zhao, Y.; Zhong, Z.; Zhao, N.; Sebe, N.; and Lee, G.~H. 2022.
\newblock Style-hallucinated dual consistency learning for domain generalized semantic segmentation.
\newblock In \emph{European Conference on Computer Vision}, 535--552. Springer.

\bibitem[{Zhou et~al.(2021)Zhou, Yang, Qiao, and Xiang}]{zhou2021domain}
Zhou, K.; Yang, Y.; Qiao, Y.; and Xiang, T. 2021.
\newblock Domain generalization with mixstyle.
\newblock \emph{arXiv preprint arXiv:2104.02008}.

\bibitem[{Zhu and Pang(2024)}]{zhu2024toward}
Zhu, J.; and Pang, G. 2024.
\newblock Toward generalist anomaly detection via in-context residual learning with few-shot sample prompts.
\newblock In \emph{Proceedings of the IEEE/CVF Conference on Computer Vision and Pattern Recognition}, 17826--17836.

\end{thebibliography}

\clearpage

\section{Appendix}
\section{Detailed Algorithm}
In this section, we present the detailed pseudo-code of the proposed FiCo for reproducibility.
\begin{algorithm}[!htb]
\caption{Pseudo-code of FiCo for one epoch training} 
\label{alg:alg1}
\begin{algorithmic}

\State \textcolor{Green}{E, $\phi$, D, C, I: teacher network, bottleneck layer, student network, DiSCo modules, DiIFi module}
\State \textcolor{Green}{$x_s$, $x_s^n$: original image, $n^{th}$ augmented view of original image}
\State \textbf{Init(E, $\phi$, D, C, I)}
\State \textbf{Opt = Adam((E, $\phi$, D, C, I).parameters())}
\For{[$x_s$, $x_s^n$] in dataloader}
  \State \textcolor{Green}{Get feature of different views from 3 blocks of the teacher network}
  \State $f^{E_1}$, $f^{E_2}$, $f^{E_3}$ = $E(x_s)$
  \State $f_n^{E_1}$, $f_n^{E_2}$, $f_n^{E_3}$ = $E(x_s^n)$
  \State \textcolor{Green}{Obtain original output from the student network}
  \State $f^{D_1}$, $f^{D_2}$, $f^{D_3}$ = $D(\phi(E(x_s)))$
    \State $f^{D_1}_n$, $f^{D_2}_n$, $f^{D_3}_n$ = $D(\phi(E(x_s^n)))$
  \State \textcolor{Green}{Compensate for distribution-specific information via DiSCo module}
  \State $f_F^{D_1}$, $f_F^{D_2}$, $f_F^{D_3}$ = $C_{1,2,3}(D(\phi(E(x_s)))) + D(\phi(E(x_s)))$
  \State $f_{F,n}^{D_1}$, $f_{F,n}^{D_2}$, $f_{F,n}^{D_3}$ = $C_{1,2,3}(D(\phi(E(x_s^n)))) + D(\phi(E(x_s^n)))$
  \State \textcolor{Green}{Filter abnormal information for invariant normality via DiIFi module}
  \State $f_{2}^{D_1}$ = $I_2(C_1(f^{D_1}))$
  \State $f_{3}^{D_1}$ = $I_3(f_{2}^{D_1})$
  \State $f_{2,n}^{D_1}$ = $I_2(C_1(f^{D_1}_n))$
  \State $f_{3,n}^{D_1}$ = $I_3(f_{2,n}^{D_1})$

  \State \textcolor{Green}{Computer the overall loss and train the whole architecture end-to-end}
  \State  $\mathcal{L}_{FiCo} = \mathcal{L}_{Fi} + \mathcal{L}_{abs} + \mathcal{L}_{Co}$
  \State $\mathcal{L}_{FiCo}$.backward()
  \State Opt.step()
\EndFor
\end{algorithmic}
\end{algorithm}

\section{Detailed Results}
\label{sec:addresults}
\quad In this section, we present the detailed results on each category for MVTec \cite{bergmann2019mvtec}, PACS \cite{li2017deeper} and CIFAR-10 \cite{krizhevsky2009learning} to expound the results presented in the main paper. 

1. Results on MVTec.
The results are shown in Table \ref{mvtec}. It can be observed that performance on most OOD scenarios has improved with a relatively large margin. 
\begin{table}[htb]
\caption{Results on MVTec for FiCo.}
\label{mvtec}
\begin{center}
\setlength{\tabcolsep}{7pt}
\begin{tabular}{cccccc}
\hline
Category & ID & Br & Co & Bl & No \\
\hline
carpet & 99.24 & 98.60 & 98.23 & 99.04 & 98.48 \\
leather & 100 & 100 & 99.93 & 100 & 100 \\
grid & 99.50 & 98.91 & 98.16 & 98.75 & 98.08 \\
tile & 99.64 & 99.93 & 100 & 99.64 & 99.60 \\
wood & 98.33 & 97.98 & 98.33 & 98.33 & 97.81 \\ 
bottle & 100 & 100 & 99.84 & 100 & 97.38 \\ 
hazelnut & 99.93 & 100 & 99.89 & 100 & 99.89 \\
cable & 96.85 & 97.73 & 96.63 & 97.4 & 96.93 \\
capsule & 98.56 & 97.01 & 94.65 & 96.77 & 86.84 \\
pill & 97 & 89.8 & 94.82 & 96.18 & 89.09 \\
transistor & 98.75 & 98.96 & 96.5 & 98.08 & 97.67 \\
metal nut & 100 & 100 & 100 & 100 & 97.85 \\
screw & 95.94 & 98.91 & 94.67 & 95.27 & 94.63 \\
toothbrush & 100 & 92.22 & 100 & 99.44 & 95.83 \\ 
zipper & 97.98 & 98.21 & 97.22 & 97.98 & 74.63 \\
\hline
Average & 98.78 & 97.88 & 97.92 & 98.46 & 94.98 \\
\hline
\end{tabular}
\end{center}
\end{table}

2. Results on PACS.
The results are shown in Table \ref{pacs}. There still exists discrepancy between ID and OOD scenarios that future research should focus on different types of domain shifts for distribution-invariant learning.

\begin{table}[htb]
\caption{Results on PACS for FiCo.}
\label{pacs}
\begin{center}
\setlength{\tabcolsep}{8pt}
\begin{tabular}{ccccc}
\hline
Category & P & A & C & S \\
\hline
dog & 83.24 & 73.03 & 59.31 & 60.15 \\
elephant & 91.56 & 70.6 & 68.89 & 73.28 \\
giraffe & 93.14 & 48.78 & 69.17 & 66.15 \\
guitar & 78.95 & 70.04 & 85.52 & 46.41 \\
horse & 82.78 & 56.89 & 64.22 & 70.28 \\
house & 99.1 & 92.68 & 91.64 & 74.89 \\
person & 99.36 & 61.09 & 57.42 & 45.07 \\
\hline
Average & 89.73 & 67.59 & 70.88 & 62.32 \\
\hline
\end{tabular}
\end{center}
\end{table}

3. Results on CIFAR-10.
The one-class novelty detection results on CIFAR-10 are shown in Table \ref{cifar10}. Gaussian noise and defocus blur are still two difficult domains that require further exploration.

\begin{table}[htb]
\caption{Results on CIFAR-10 for FiCo.}
\label{cifar10}
\begin{center}
\setlength{\tabcolsep}{7pt}
\begin{tabular}{cccccc}
\hline
Category & ID & Br & Co & Bl & No \\
\hline
airplane & 83.47 & 80.33 & 74.65 & 70.73 & 67.67 \\
automobile & 88.91 & 86.24 & 69.9 & 63.02 & 75.22 \\
bird & 74.79 & 72.54 & 64.1 & 58.11 & 53.75 \\
cat & 56.31 & 53.08 & 55.81 & 52.53 & 52.25 \\
deer & 79.11 & 78.39 & 58.1 & 63.85 & 51.72 \\
dog & 73.19 & 67.99 & 63.77 & 53.15 & 63.19 \\
frog & 83.84 & 81.09 & 69.8 & 62.69 & 61.87 \\
horse & 86.54 & 83.97 & 76.51 & 69.18 & 69.26 \\
ship & 89.82 & 88.25 & 81.85 & 76.71 & 73.23 \\
truck & 88.87 & 86.29 & 77.85 & 67.57 & 75.9 \\
\hline
Average & 80.49 & 77.82 & 69.23 & 63.75 & 64.41\\
\hline
\end{tabular}
\end{center}
\end{table}

\end{document}